# Integrated Data Analysis of Plasma Electron Density Profile Tomography for HL-3 with Gaussian Process Regression

Cong Wang[1,+], Jiahong Chen[3], Renjie Yang[1], Dong Li[2], Zhibin Wang[3], Zongyu Yang[2], Zhijun Wang[1], Yixiong Wei[1], Zhaoyang Liu[1], Chenshu Hu[1], Jing Li[1,*]

*1 Zhejiang Lab, Hangzhou 310000, China*

*2 Southwestern Institute of Physics, Chengdu 610041, China*

*3 Sino-French Institute of Nuclear Engineering and Technology, Sun Yat-sen University, Zhuhai 519082, China*

**Corresponding author*

*+Contact person*

## Abstract

An integrated data analysis model based on Gaussian Process Regression is proposed for plasma electron density profile tomography in the HL-3 tokamak. The model combines line-integral measurements from the far-infrared laser interferometer with point measurements obtained via the frequency-modulated continuous wave reflectometry. By employing Gaussian Process Regression, the model effectively incorporates point measurements into 2D profile reconstructions, while coordinate mapping integrates magnetic equilibrium information. The average relative error of the reconstructed profile obtained by the integrated data analysis model with normalized magnetic flux is as low as $3.60 \times 10^{-4}$. Additionally, sensitivity tests were conducted on the grid resolution, the standard deviation of diagnostic data, and noise levels, providing a robust foundation for the real application to experimental data.

**Keywords:** Bayesian tomography; Gaussian Process; HL-3; 2D profile; Plasma electron density profile

## 1 Introduction

In nuclear fusion research, analysis techniques are crucial for the inference of physical quantities from the measured data. Currently,

numerous diagnostic systems have been developed to obtain information about a wide range of physical parameters. Based on these diagnostic results, studies on magnetohydrodynamic instability (MHD), turbulence-related anomalous transport and other physical phenomena are conducted. Traditional diagnostic data analysis typically relies on a single diagnostic system. Data from individual diagnostics are analyzed separately using various techniques, such as forward modeling tools like $\chi^2$ fitting or backward inversion methods like Abel inversion[1,2]. However, these methods often suffer from inconsistency and can be time-cconsuming[3], hindering the understanding of physics of plasma.

In current and next-generation nuclear fusion experiments, it is crucial to develop analysis techniques that integrate diverse diagnostics while efficiently utilizing information from various data sources. This integrated approach enables consistent analysis across multiple diagnostics and facilitates the extraction of significant insights from heterogeneous data[4]. Integrated Data Analysis (IDA), based on Bayesian probability theory, has emerged as a primary research direction[5,6]. Notably, MINERVA, developed by J. Svensson and A. Werner, is the most widely used IDA platform within the Bayesian probability theory (BPT) framework, employed in various fusion devices such as JET[7,8], ASDEX-Upgrade[3,9], MAST[10,11], HL-2A[12] and W7-X[13,14]. Within this framework, researchers implement integrated analyses of multiple sources of diagnostic data, leveraging BPT to make informed decisions and extract meaningful insights[15]. Extensive Bayesian research has been conducted on different fusion devices with promising results[16–24]. In the Bayesian inference framework, developing a forward model of a diagnostic system is essential and requires a comprehensive understanding of the diagnostic principles[25]. Sehyun Kwak et al.[23] first calculated the posterior distribution of $n_e$ and $T_e$ given the $D_{int}$ and $D_{TS}$ using forward models of the interferometer and Thomson scattering system. Subsequently, the posterior distribution of $Z_{eff}$ is calculated, given $D_{vis}$ and $D_{IR}$, with the forward models of the visible and near-infrared spectrometers[22].

In precious research, Gaussian Process Regression (GPR) was applied for 1D profile inference. Chilenski, M.A. et al. applied GPR for plasma profile inference[26], demonstrating its ability to fit smooth curves to noise, discrete observations of plasma profiles with the mapped midplane major radius. T. Nishizawa et al. introduced an alternative GPR approach to estimate plasma parameter profiles and their derivatives[27]. This method provides meaningful measurements of the electron density profile and its derivative based on arbitrary linear observations, considering finite spatial resolution of diagnostics with a sensitivity matrix. These studies all employ GPR for 1D profile fitting, utilizing the mapped midplane major radius or the locations of magnetic flux surfaces obtained from equilibrium reconstruction[28–30]. Consequently, the measurement position information in 2D space is replaced by the locations of 1D magnetic flux surfaces, limiting the inversion to 1D.

To overcome the dimensional limitations of existing approaches, we propose an IDA model based on Gaussian Process Regression (GPR) for electron density profile tomography. Building upon the previous IDA implementations for HL-2A[12,15,16,31,32] , our model combines point measurements from frequency modulated continuous wave reflectometry (FMCW) with line-integrated data from Far-infrared interferometry (FIR) within the Bayesian framework. It employs GPR to incorporate the spatial positions and point measurements from FMCW into the 2D reconstruction framework. This approach offers two key advantages: (1) it eliminates the need for establishing a corresponding forward model for FMCW, and (2) it preserves the inherently 2D nature of the inversion results. The spatial coordinates of point measurements are determined by magnetic flux surfaces at the plasma boundaries. Furthermore, the sensitivity matrix[27] is used to incorporate magnetic equilibrium information under the 2D analysis framework through coordinate mapping.

The structure of this paper is as follows: **Section 2** introduces the theory of the IDA model. In **Section 3**, the IDA model is validated. **Section 4** presents the sensitivity analysis of the IDA model. Finally, the conclusions are given in **Section 5**.

# 2 Model

Bayesian probability theory is briefly introduced in Appendix A. $F$ represents the profile parameters of interest and $v_*$ denotes the observed point values. In this study, a probabilistic model is employed to compute the posterior distribution $P(F|d, v_*; \theta)$, as depicted in Eq.(2-1), by integrating line-integral measurements $d$ and point values $v_*$, where $\theta$ represents the model parameters. It is important to note that the point values $v_*$ consist of both the inversion results $F_*$ obtained from a conventional algorithm and an error term $\varepsilon_*$, such that $v_* = F_* + \varepsilon_*$.

$$P(F|d, v_*; \theta) \propto P(d|F, v_*; \theta) \cdot P(F|v_*; \theta) \tag{2-1}$$

The IDA model includes point values $v_*$ and line-integral measurements $d$ through two sequential stages. First, the conditional probability distribution $P(F|v_*; \theta)$ is derived by integrating the point values $v_*$ into the prior distribution $P(F|\theta)$ using GPR. Second, the line-integral measurements are integrated into the Bayesian model by formulating the likelihood function $P(d|F, v_*; \theta)$. This is achieved through the forward modeling of the line measurements diagnostic, which complies with methodologies employed in previous studies[16,22,24,33]. By combining the likelihood function with the joint probability distribution, the posterior distribution $P(F|d, v_*; \theta)$ is inferred. In this section, all the parameter definitions and theoretical derivations adhere to the conventions established in the classic textbook[34] and the tutorial[35,36].

## 2.1 Conditional probability distribution by GPR

We assume that $F$ can be represented by a Gaussian filed characterized by a mean function $m(\cdot)$ and a kernel function $k(\cdot,\cdot)$. For $F$ evaluated at locations $[F(x_1), F(x_2), \cdots, F(x_g)]$, it can be modeled as a multivariate Gaussian distribution, expressed as:

$$[F(x_1), F(x_2), \cdots, F(x_g)] \sim \mathcal{N}(\mu, \Sigma), \quad (2\text{-}2)$$

where $\mu = [m(x_1), \ldots m(x_g)]$ is the mean vector, $g$ is the number of all locations, and the covariance matrix $\Sigma$ is defined by the kernel function such that $\Sigma_{j,k} = k(x_j, x_k)$, whose definition would be discussed in **Section 2.3**. Without loss of generality, the mean function is assumed to be 0, i.e., $m(\cdot) = 0$, in the absence of any prior information.

Given observed data $F_* = [F_{*1}, F_{*2}, \cdots, F_{*i}]$ at points $x_* = [x_{*1}, x_{*2}, \cdots, x_{*i}]$, the joint distribution of $F$ at point $x$ with $F_*$ can be expressed as:

$$\begin{bmatrix} F_*^T \\ F^T \end{bmatrix} \sim \mathcal{N}\left(\begin{bmatrix} 0 \\ 0 \end{bmatrix}, \begin{bmatrix} \Sigma_{**} & \Sigma_* \\ \Sigma_*^T & \Sigma \end{bmatrix}\right), \quad (2\text{-}3)$$

where $\Sigma_{**j,k} = k(x_{*j}, x_{*k})$, and $\Sigma_{*j,k} = k(x_{*j}, x_k)$. The conditional distribution of $F$ given $F_*$ can be derived as:

$$F|F_*; \theta \sim \mathcal{N}(\Sigma_*^T \Sigma_{**}^{-1} F_*, \Sigma - \Sigma_*^T \Sigma_{**}^{-1} \Sigma_*), \quad (2\text{-}4)$$

In practice, the observed point values $v_*$ are composed of both the inversion results $F_*$ from a conventional algorithm and an error term $\varepsilon_*$, such that, $v_* = F_* + \varepsilon_*$. Typically, $\varepsilon_*$ is assumed to be identically and independently distributed (i.i.d.) Gaussian noise with a zero mean and variance $\sigma_*^2$. The conditional probability distribution is then given by Eq.(2-5) and abbreviated as $\mathcal{N}(\mu_1, \Sigma_1)$:

$$\begin{gathered} F|v_*; \theta \sim \mathcal{N}(\Sigma_*^T (\Sigma_{**} + \sigma_*^2 I)^{-1} v_*,\ \Sigma - \Sigma_*^T (\Sigma_{**} + \sigma_*^2 I)^{-1} \Sigma_*), \\ \mu_1 = \Sigma_*^T (\Sigma_{**} + \sigma_*^2 I)^{-1} v_*,\ \Sigma_1 = \Sigma - \Sigma_*^T (\Sigma_{**} + \sigma_*^2 I)^{-1} \Sigma_*. \end{gathered} \quad (2\text{-}5)$$

## 2.2 Likelihood by forward model

The line integral measurements obtained from diagnostics can be characterized using the forward model, along with a systematic and unified error analysis. This is demonstrated in Eq.(2-6), which describes the relationship between the forward model and the comprehensive error:

$$d = G(F) + \varepsilon, \tag{2-6}$$

where $G$ corresponds to the forward model, and $\epsilon$ represents the noise associated with the observations. For a linear relationship, the expression $G(F)$ can be represented as $R$ multiplied by $F$[32,37], where $R$ is the response matrix[38], which denotes the path length of the line of sights. In most experimental scenarios, the noise follows an independent Gaussian distribution with mean of 0.

For the likelihood function $P(d|F;\theta)$, it quantifies how well the model, given the measured signal $d$, fits the predicted signal obtained from the forward model $G(F)$. The measurement noise $\varepsilon = d - R \cdot F$ in Eq. (2-6) is assumed to be an i.i.d. Gaussian noise with a mean of 0 and a covariance matrix $\Sigma_\varepsilon$. Here, $\Sigma_\varepsilon$ is a diagonal matrix whose entries define the data variance based on an error analysis of the measured data in actual experiments. As a result, the probability density function (PDF) $p(d|F, v_*; \theta)$ can be expressed as Eq. (2-7), where $n$ is the dimension of $d$:

$$p(d|F, v_*; \theta) = \frac{1}{(2\pi)^{\frac{n}{2}} |\Sigma_\varepsilon|^{\frac{1}{2}}} exp\left[-\frac{1}{2}(d - R \cdot F)^T \Sigma_\varepsilon{}^{-1}(d - R \cdot F)\right] \tag{2-7}$$

The PDF $p(F|v_*; \theta)$ can be expressed as Eq. (2-8):

$$p(F|v_*; \theta) = \frac{1}{(2\pi)^{\frac{g}{2}} |\Sigma_1|^{\frac{1}{2}}} exp\left[-\frac{1}{2}(F - \mu_1)^T \Sigma_1^{-1}(F - \mu_1)\right], \tag{2-8}$$

where $\Sigma_1$ and $\mu_1$ are shown in Eq.(2-5). Finally, the posterior density function is given by:

$$\begin{aligned} p(F|d, v_*; \theta) &\propto p(d|F, v_*; \theta) \cdot p(F|v_*; \theta) \\ &\propto \frac{1}{(2\pi)^{\frac{n}{2}} |\Sigma_\varepsilon|^{\frac{1}{2}}} exp\left[-\frac{1}{2}(d - R \cdot F)^T \Sigma_\varepsilon^{-1}(d - R \cdot F)\right] \cdot \\ &\quad \frac{1}{(2\pi)^{\frac{g}{2}} |\Sigma_1|^{\frac{1}{2}}} exp\left[-\frac{1}{2}(F - \mu_1)^T \Sigma_1^{-1}(F - \mu_1)\right]. \end{aligned} \tag{2-9}$$

Hence, the posterior mean vector and covariance matrix can be given by

$$\mu^{post} = \mu_1 + (R^T \Sigma_\varepsilon^{-1} R + \Sigma_1^{-1})^{-1} R^T \Sigma_\varepsilon^{-1} (d - R \cdot \mu_1), \tag{2-10}$$

$$\Sigma^{post} = \left(R^T {\Sigma_\varepsilon}^{-1} R + \Sigma_1^{-1}\right)^{-1}. \tag{2-11}$$

## 2.3 Kernel functions for the Gaussian Process prior

Different kernel functions can be chosen to derive the covariance matrices $(\Sigma, \Sigma_*, \Sigma_{**})$ above. Due to the characteristics of the Squared Exponential (SE) kernel function[31,37,39], as depicted in Eq.(2-12), it is utilized in this work.

$$k_{jk}^{SE}(x_j, x_k) = \sigma^2 \, exp\left(-\frac{dist^2}{2l^2}\right), dist = \left\|x_j - x_k\right\| \tag{2-12}$$

$\sigma$ is the scale factor that determines the magnitude of the random field; $dist$ denotes the distance between any pair of positions $x_j$ and $x_k$, which is scaled by the correlation length $l$. Here, the position corresponds to the Cartesian coordinates of the profile.

Generally, some profiles are considered functions of the normalized magnetic flux ($\boldsymbol{\psi}$). For instance, the electron density profile on the same magnetic surface is uniform. Based on this assumption, solving the 2D profile ($F(\boldsymbol{x})$) can be reduced to inverting a 1D profile ($F(\boldsymbol{\psi})$). This is achieved by mapping the 2D Cartesian coordinates $\boldsymbol{x} = [x_1, \dots x_g]$ to $\boldsymbol{\psi} = [\psi_1, \dots \psi_g]$, which is obtained using EFITNN[28]. The transformed SE kernel function retains the 2D tomography characteristics by mapping the 2D distances $\left\|x_j - x_k\right\|$ into 1D distances $|\psi_j - \psi_k|$, as presented in Eq. (2-13).

$$k_{jk}^{SE}(\psi_j, \psi_k) = \sigma^2 \, exp\left(-\frac{dist^2}{2l^2}\right), dist = |\psi_j - \psi_k| \tag{2-13}$$

# 3 Model validation

In this study, the FMCW diagnostic system and FIR diagnostic system of the HL-3 Tokamak, depicted in **Figure 1**, are employed to achieve the electron density profile tomography inside the Last Closed Flux Surface (LCFS). Detail information about these diagnostics is provided in Appendix B.

In this section, a known 2D electron density profile, as shown in **Figure 2**(a), is utilized to evaluate the reconstructed profile. Detail information on constructing the 2D electron density profile is introduced in Appendix C. When the plasma electron density profile is known, the corresponding synthetic diagnostics data can be obtained through the forward model. Here, the synthetic diagnostic data from the virtual diagnostic of FIR serves as the input $d$. The synthetic values from the virtual diagnostic of FMCW on the electron density profile are obtained as the input $v_*$, represented by the colored points shown in **Figure 2**(b). Systematic errors of $v_*$ introduced by EFIT equilibrium reconstruction, are ignored in this work. The standard deviations of $d$ and $v_*$ are set to 0. The accuracy of the model can then be evaluated by comparing the reconstructed profile $F$ with the synthetic profile $F_p$.

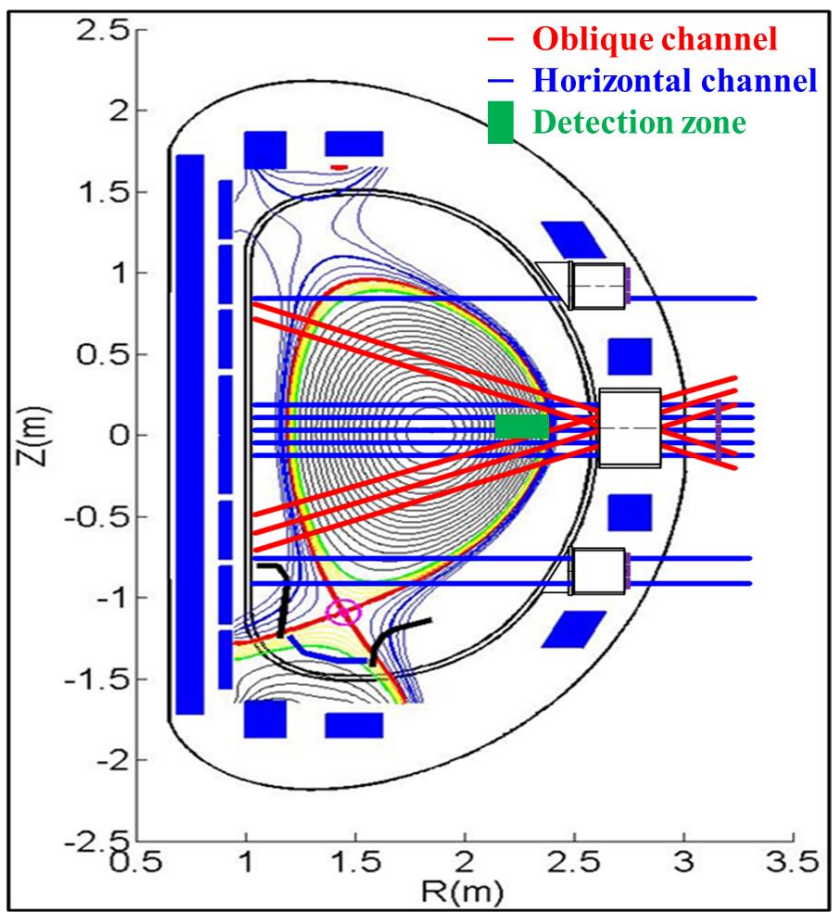

**Figure 1** Poloidal view of the typical plasma configuration on HL-3 with two primary density diagnostics mapped to the same cross-section (R is major radius; Z is the axis of axisymmetry located at R = 0 m):

detection zone of the FMCW diagnostic system (green area) and 13 detection channels of FIR diagnostic system (blue lines and red lines).

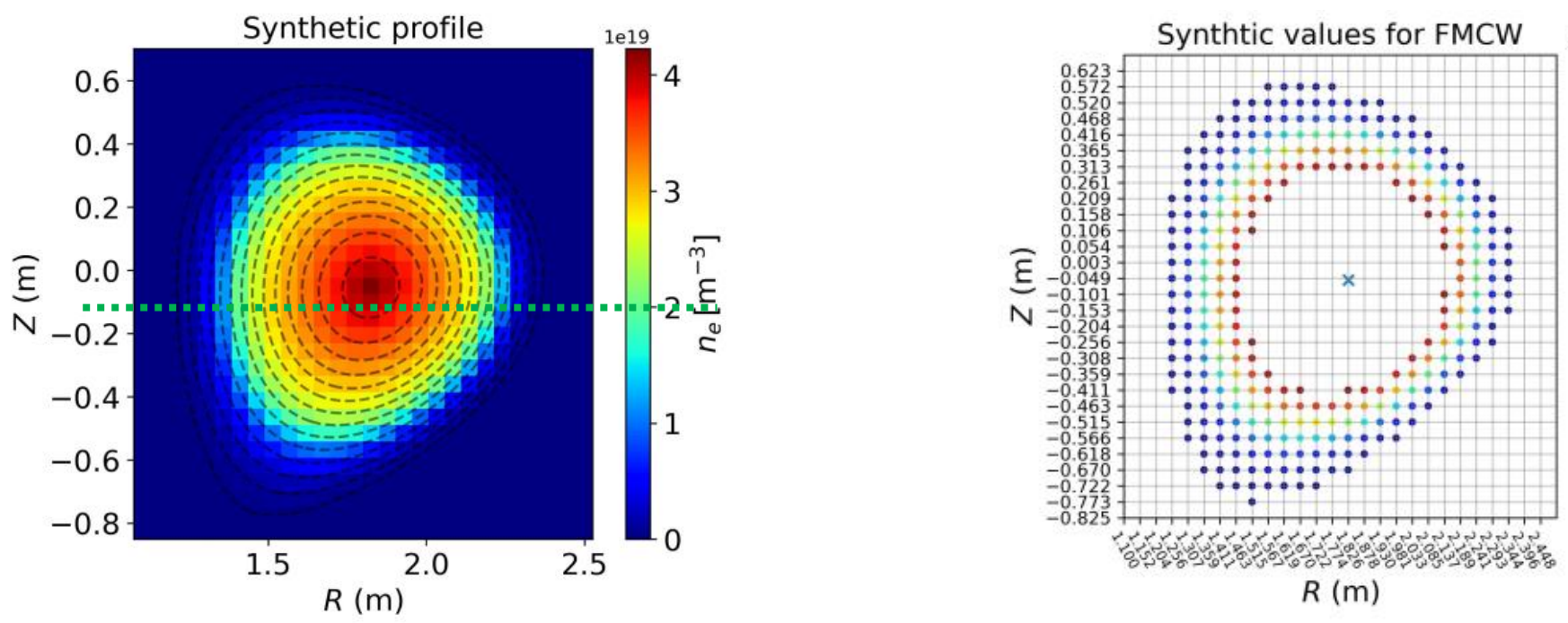

(a) (b)

**Figure 2** (a) synthetic profile of electron density with green dot line at middle of Z plane; (b) synthetic values form FMCW.

Two indicators, relative root-mean-square error (rRMSE) and average relative error ($\bar{\xi}$), are used to evaluate the results[18]. The rRMSE measures the overall deviation between the reconstructed and synthetic profiles, capturing the discrepancy between the two sets of values. The vector $\boldsymbol{\xi} = [\xi_1, \dots \xi_g]$ assesses the relative errors for all nodes, indicating the difference between the reconstructed value and the synthetic value (true value). $\bar{\xi}$ represents the average of $\boldsymbol{\xi}$ and $\xi_{\max}$ donates the maximum value of $\boldsymbol{\xi}$. They are expressed as:

$$rRMSE = \sqrt{\frac{\frac{1}{g}\sum_{i=1}^{g}\left(F^i - F_p^i\right)^2}{\sum_{i=1}^{g}\left(F_p^i\right)^2}} \tag{3-1}$$

$$\xi_i = \frac{\left|F^i - F_p^i\right|}{F_p^{max}} \tag{3-2}$$

$$\bar{\xi} = \frac{\sum_{i=1}^{g}\xi^i}{g}, \xi_{max} = max(\boldsymbol{\xi}) \tag{3-3}$$

, where $F^i$ is the reconstructed value at $i$-th node, $F_p^i$ is the synthetic value at $i$-th node, and $F_p^{max}$ is the maximum value of the synthetic profile.

We compared the results from Bayesian data analysis (BDA) model with IDA model using different kernel functions. In the BDA model, only the line-integrated measurement $d$ is analyzed using the method presented in previous work[15,31,33,38,40]. Based on the Bayesian probability theory, the mean vector and covariance matrix of the posterior distribution can be obtained analogously from Eq.(2-10) and Eq.(2-11) as

$$\mu_F^{post} = \mu_F' + \left(R^T \Sigma_\varepsilon{}^{-1} R + \Sigma_F{}'^{-1}\right)^{-1} R^T \Sigma_\varepsilon{}^{-1} (d - R \cdot \mu_F'), \quad (3\text{-}4)$$

$$\Sigma_F^{post} = \left(R^T \Sigma_\varepsilon{}^{-1} R + \Sigma_F{}'^{-1}\right)^{-1}. \quad (3\text{-}5)$$

The prior probability $p(F|I)$ is a Gaussian distribution with a zero mean $\mu_F{}'$. The covariance matrix $\Sigma_F{}'$ is calculated by Eq. (2-12) and further modified that its value outside the LCFS is set to be a small value. This modification indirectly introduces LCFS information, imposes strong constraints on $\Sigma_F{}'$, and results in more plausible outcomes.

### 3.1 Synthetic test with $k_{jk}^{SE}\left(x_j, x_k\right)$

The inversion results using $k_{jk}^{SE}\left(x_j, x_k\right)$, including reconstructed profiles, reconstructed values along the Z middle plane and $\boldsymbol{\xi}$ from both BDA model and IDA model, are illustrated in **Figure 3**.

From **Figure 3**(a) and (c), the reconstructed value from the BDA model appears as a rather smooth curve due to the large length scale of the SE kernel function. However, the reconstructed values at the edge and core of the plasma, as seen in the locally enlarged area in **Figure 3**(c), are inconsistent with the synthetic profile. For the BDA model, it is challenging to accurately capture the values during model inversion using only line-integral measurements from FIR and the SE kernel function. Consequently, the reconstructed profile fails to adequately reflect the varying gradient of the synthetic profile, leading to large relative errors in areas of steep gradients, as shown in **Figure 3**(e).

In contrast, **Figure 3**(b) shows that the reconstructed profile from the IDA model which is inferior to that of the BDA model, with distortions evident in the core region. The reason for distortions is the scarcity of measurement channels in the core, which leads to inaccurate inversion results. As depicted in **Figure 3**(d), the results in the core area have large uncertainty. This is because the standard deviation of $v_*$ is set to zero, which reduces the uncertainty of the inversion results in the detection zone (green area in **Figure 1**) at the edges, thereby increasing the uncertainty in the core area. It is therefore projected that with the incorporation of additional diagnostic measurement signals and a more strategic diagnostic measurement layout, IDA model could achieve superior outcomes in the core region. This improvement would potentially decouple the IDA-type approach from magnetic equilibrium, enabling the verification of magnetic equilibrium in the core of plasma from inversion results. It is worth mentioning that as indicated by the red dashed box areas in **Figure 3**(d), the reconstructed and synthetic profiles agree well in the ranges of 2.2-2.4 m and 1.2-1.4 m. This demonstrates the effectiveness of GPR in constraining point measurements as expected.

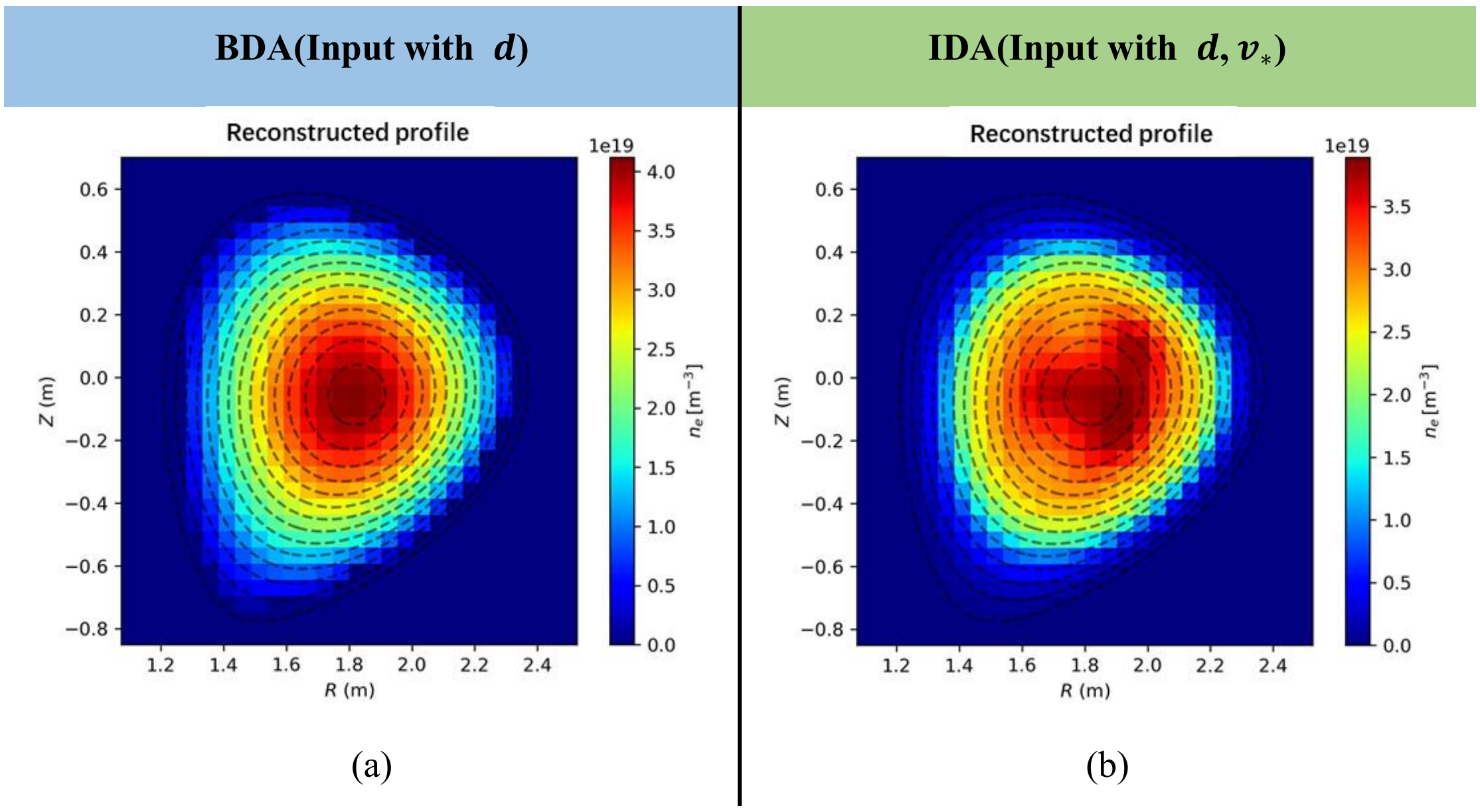

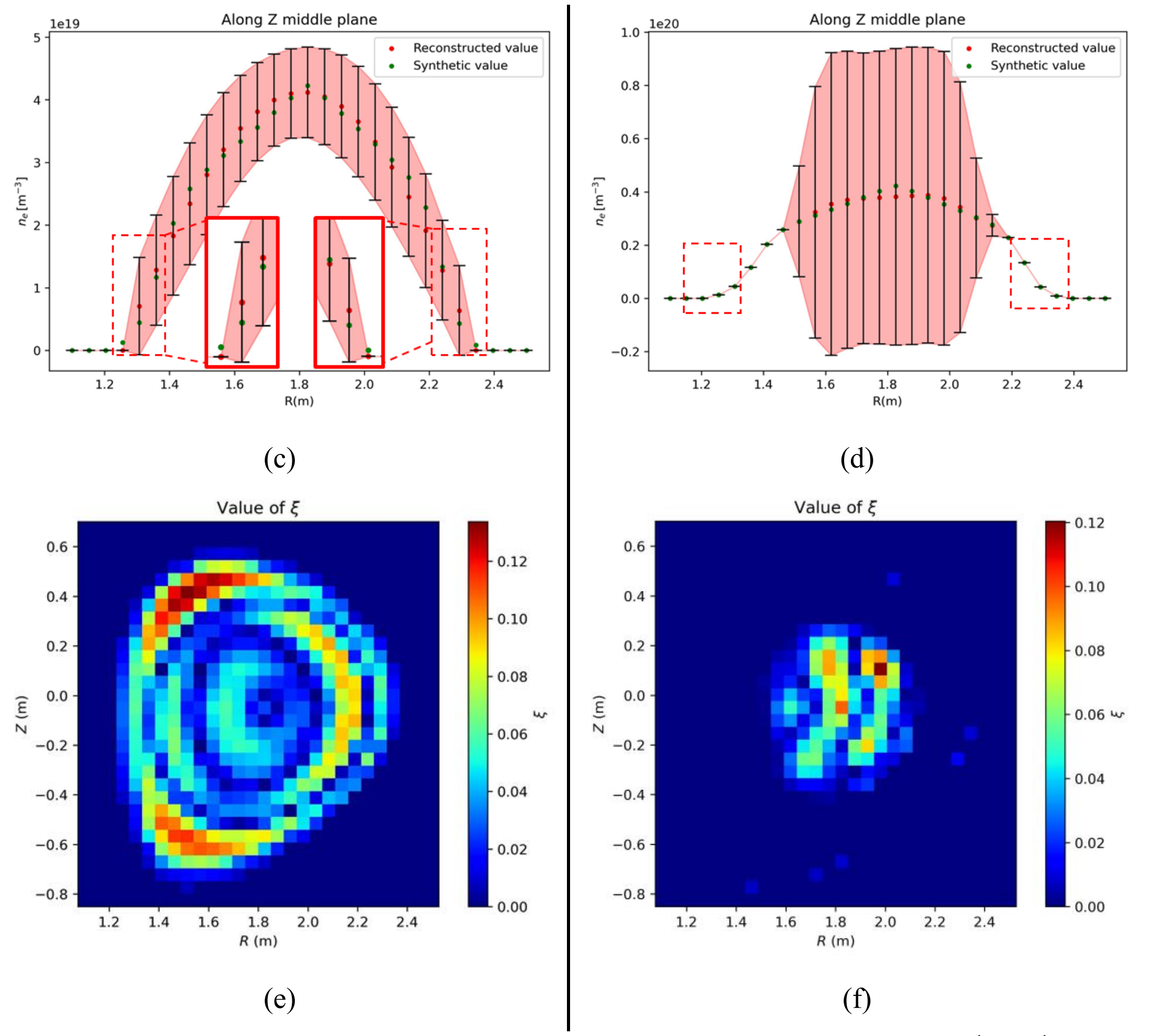

**Figure 3** The inversion results of BDA model (left) and IDA model (right) utilizing $k_{jk}^{SE}(x_j, x_k)$. (a) and (b) depict the reconstructed profiles. The black dotted line in both panels represents the self-consistent equilibrium magnetic flux, as mentioned in **Figure 11** of Appendix C. (c) and (d) illustrate the reconstructed values along the Z midplane (green dotted line in **Figure 2**(a)). The green dots indicate the true plasma electron density in the synthetic profile, the red dots represent the reconstructed plasma electron density, and the red shaded area indicates the uncertainty of the reconstructed value. (e) and (f) show the relative error ξ at each grid node.

**Table 1** summarizes the evaluation parameters of the different models. From the perspective of evaluation parameters, the IDA model produces a more precise electron density profile compared to the BDA model. For the BDA model, $\bar{\xi}$ is $2.25 \times 10^{-2}$ and $\xi_{max}$, which occurs at the edge of the plasma (see **Figure 3**(e)), is $1.34 \times 10^{-1}$. In contrast, for the IDA model, $\bar{\xi}$ is $5.18 \times 10^{-3}$ and $\xi_{max}$, which occurs at the core of the plasma (see **Figure 3**(f)), is $1.20 \times 10^{-1}$. The rRMSE for the BDA model is $3.55 \times 10^{-3}$, whereas for the IDA model is $1.56 \times 10^{-3}$.

**Table 1** Comparison of evaluation parameters of different models with $k_{jk}^{SE}(x_j, x_k)$.

| Model | $\xi_{max}$ | $\bar{\xi}$ | rRMSE |
|---|---|---|---|
| BDA | $1.34 \times 10^{-1}$ | $2.25 \times 10^{-2}$ | $3.55 \times 10^{-3}$ |
| IDA | $\mathbf{1.20 \times 10^{-1}}$ | $\mathbf{5.18 \times 10^{-3}}$ | $\mathbf{1.56 \times 10^{-3}}$ |

## 3.2 Synthetic test with $k_{jk}^{SE}(\psi_j, \psi_k)$

To retain the characteristics of 2D tomography and introduce the normalized magnetic flux surface $\boldsymbol{\psi}$, the kernel $k_{jk}^{SE}(\psi_j, \psi_k)$ is applied. For each coordinate $x$, there is a corresponding normalized magnetic flux $\boldsymbol{\psi}$ in the low-dimensional space. The inversion results are shown in **Figure 4**.

By incorporating magnetic equilibrium information, both models demonstrate improved performance relative to the results presented in **Section 3.1.** Specifically, the BDA model achieves enhanced reconstruction accuracy in both edge and core plasma regions (**Figure 4**(c)), with the reconstructed profile now closely matching the sharp gradient of the synthetic profile. Furthermore, the uncertainty in reconstructed values is significantly reduced compared to **Figure 3**(c), demonstrating the constraining effect of magnetic equilibrium information.

**Figure 4**(d) and (f) show that the reconstructed results of the IDA model fit well in both the core and edge regions. Given that GPR constrains the point values at the edges, the error associated with the reconstructed values at the plasma edge is minimized. Furthermore, by incorporating normalized magnetic flux, the IDA model's performance in reconstructing the core of the profile is enhanced. The uncertainty in the core region arises from the absence of data to constrain predictions is reduced. The reconstructed values show better consistency with the synthetic fields than those in **Figure 3**(d).

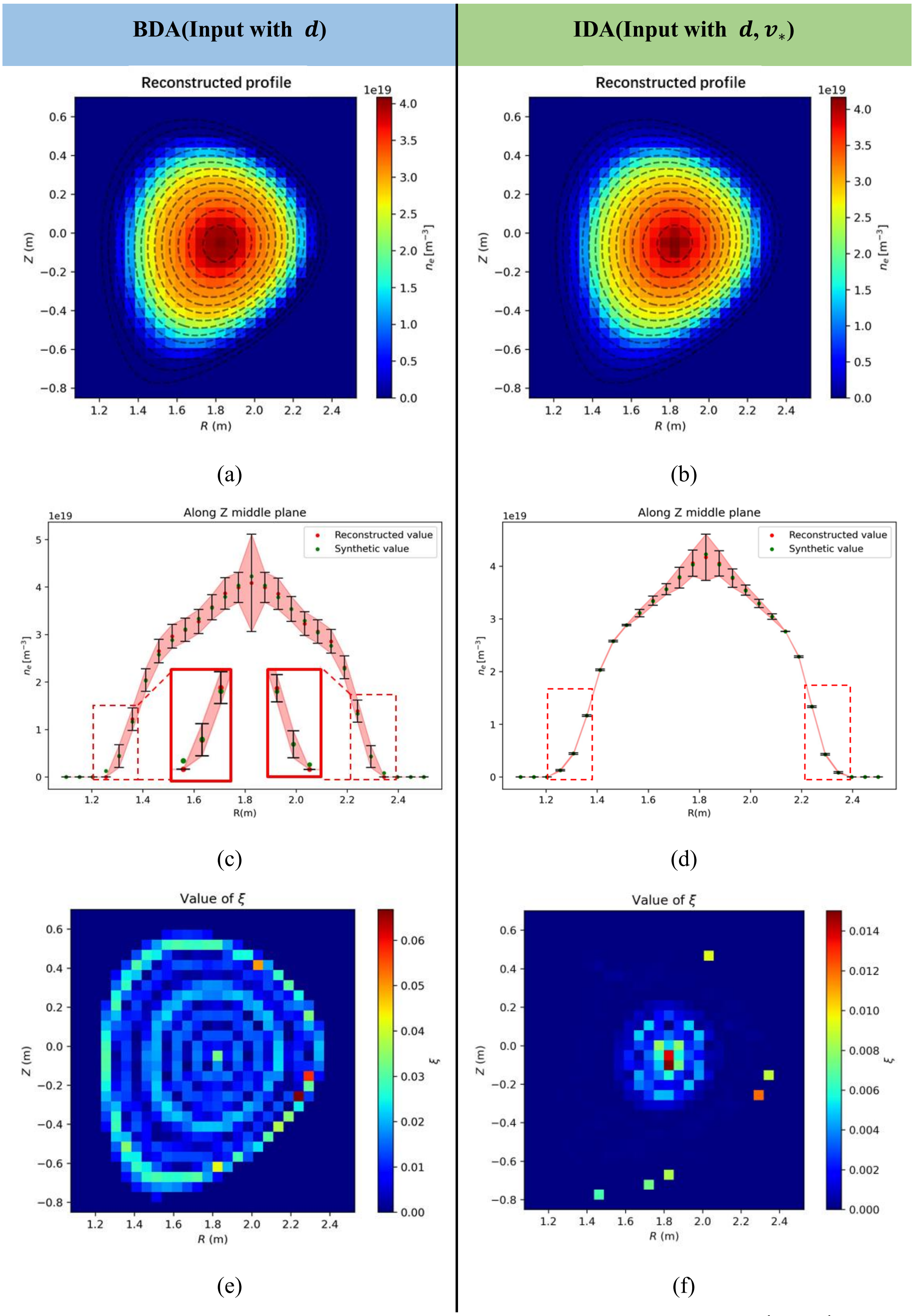

**Figure 4** The inversion results of BDA model (left) and IDA model (right) using $k_{jk}^{SE}(\psi_j, \psi_k)$. (a) and (b) depict the reconstructed profiles. The dotted line in both panels represents the self-consistent equilibrium magnetic flux, as mentioned in **Figure 11** of Appendix C. (c) and (d) illustrate the reconstructed values along Z middle plane (green dot line in **Figure 2**(a)). The green dots indicate the true plasma electron density in the synthetic profile, the red dots represent the reconstructed plasma

electron density, and the red shaded area indicates the uncertainty of the reconstructed value. (e) and (f) show the relative error ξ at each grid node.

**Table 2** presents a comparison of the evaluation parameters for different models. For the BDA model, the $\bar{\xi}$ decreases to $6.68 \times 10^{-3}$ and the $\xi_{max}$ is $6.68 \times 10^{-2}$. For the IDA model, the $\bar{\xi}$ decreases to $3.60 \times 10^{-4}$ and the $\xi_{max}$ is $1.50 \times 10^{-2}$. The minimum rRMSE is $1.32 \times 10^{-4}$. These results demonstrate a direct correlation between the accuracy of prior information in the Bayesian framework and reconstruction precision. The improved profile reconstruction verifies both the robustness and scalability of the IDA modeling approach. The IDA model with $k_{jk}^{SE}(\psi_j, \psi_k)$ will be utilized in the subsequent analysis.

**Table 2** Comparison of evaluation parameters of different models with $k_{jk}^{SE}(\psi_j, \psi_k)$.

| Input | $\xi_{max}$ | $\bar{\xi}$ | rRMSE |
|---|---|---|---|
| BDA | $6.68 \times 10^{-2}$ | $6.68 \times 10^{-3}$ | $1.06 \times 10^{-3}$ |
| IDA | $\mathbf{1.50 \times 10^{-2}}$ | $\mathbf{3.60 \times 10^{-4}}$ | $\mathbf{1.32 \times 10^{-4}}$ |

# 4 Model sensitivity analysis

## 4.1 Mesh sensitivity analysis

To test grid sensitivity and determine the optimal resolution of the profile suitable for the IDA model with $k_{jk}^{SE}(\psi_j, \psi_k)$, we use four different grid sizes: 210 (14×15), 840 (28×30), 1890 (42×45), and 3360 (56×60). The aspect ratio of the R to Z dimensions for the profile is based on the inversion region of HL-3, set at 14:15. Accordingly, the number of grids in the R and Z directions is adjusted to maintain each grid cell's shape as close to square as possible. This section constructs four synthetic profiles of electron density with varying grid numbers, as shown in **Figure 5**. The standard deviations of $d$ and $v_*$ are set to 0.

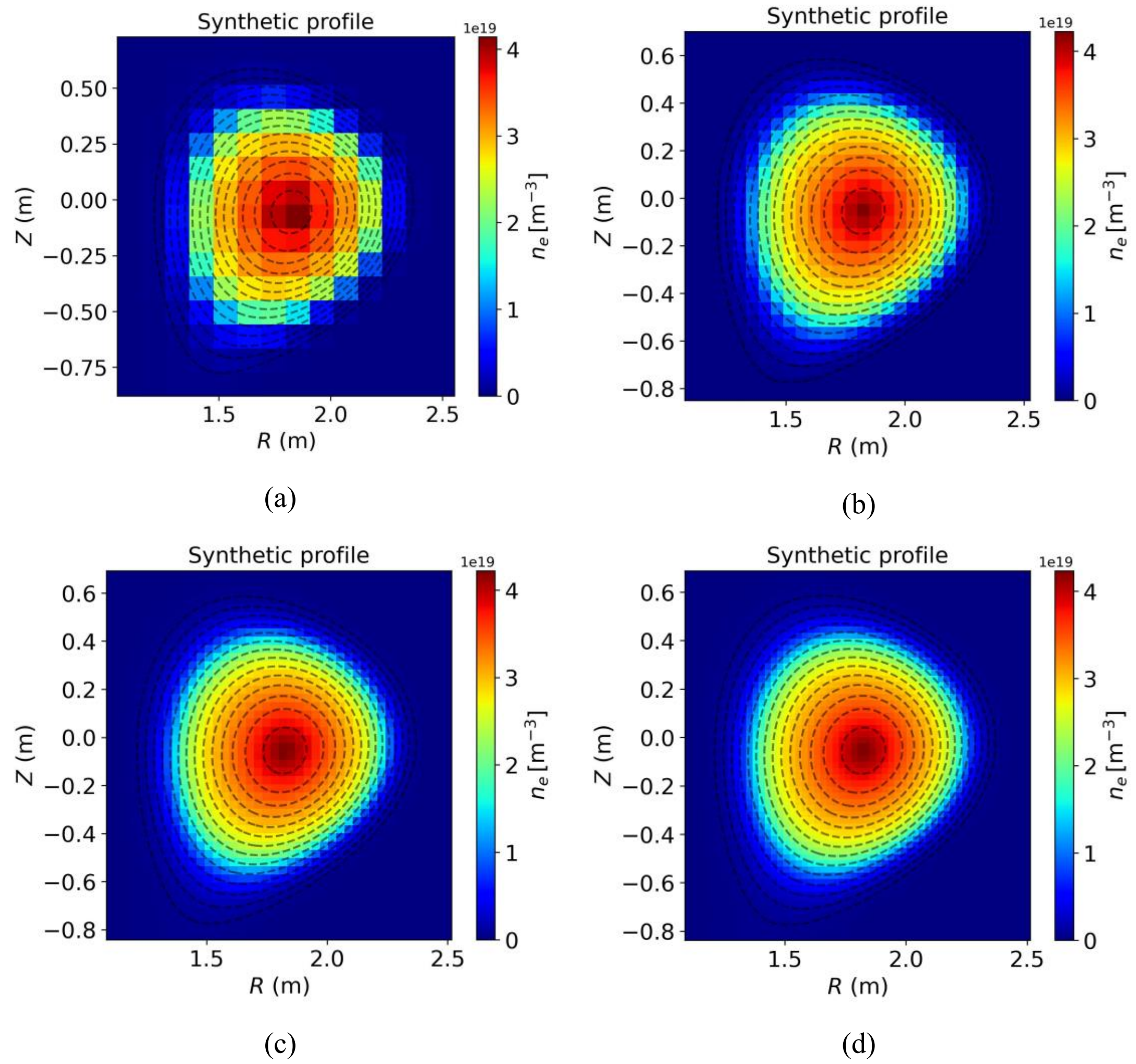

**Figure 5** Synthetic electron density profiles with different numbers of grids. (a) The synthetic profile with 14×15 grids. (b) The synthetic profile with 28×30 grids. (c) The synthetic profile with 42×45 grids. (d) The synthetic profile with 56×60 grids.

**Figure 6** presents the inversion results for the IDA model with varying grid numbers. Column (a) indicates that as the number of grids increases, the resolution of the 2D reconstructed profile improves, resulting in a smoother result. However, when the grid numbers reach 1890 (42×45) and 3360 (56×60), small perturbations appear in the core of the reconstructed profiles. As depicted in column (b), the uncertainty associated with these perturbations is notably high. This phenomenon may be preliminarily attributed to instabilities arising from the numerical calculations. **Table 3** provides a comparative overview of the evaluation parameters for all four cases examined. The IDA model using 840 (28×30) grid points exhibits the best performance among the four cases evaluated. The three metrics— $\xi_{max}$, $\bar{\xi}$, and rRMSE—are all minimized in this configuration: $\xi_{max} = 1.50 \times 10^{-2}$, $\bar{\xi} = 3.60 \times 10^{-4}$, and rRMSE=$1.32 \times 10^{-4}$. Additionally, the inversion results obtained with this grid resolution do not exhibit any signs of numerical instability.

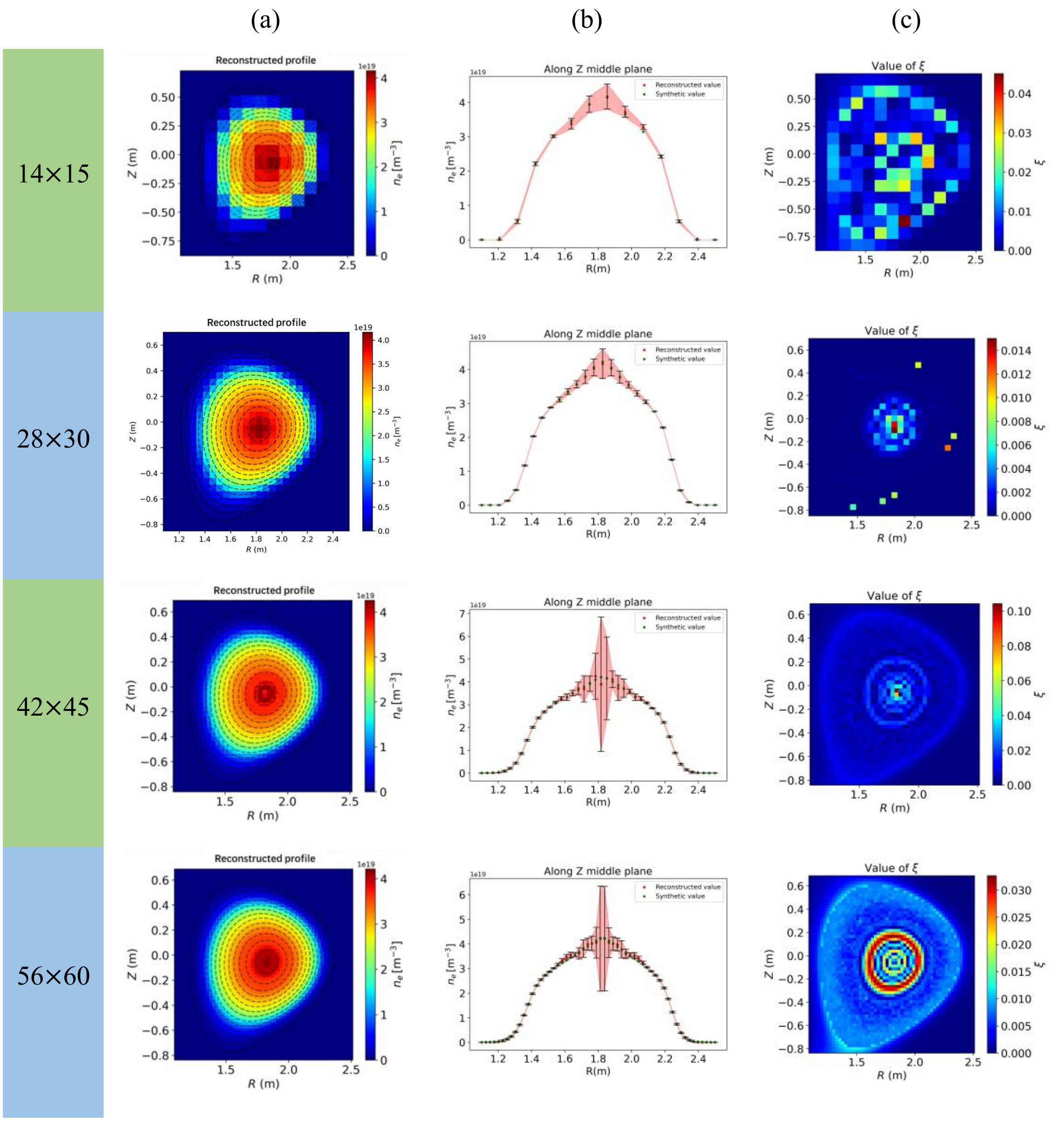

**Figure 6** The inversion results for the IDA model with varying grid numbers. Column (a): the reconstructed profiles. Column (b): the plasma electron density along the Z midplane of the reconstructed profiles. Column (c): the relative error ξ at each grid node.

**Table 3** Comparison of evaluation parameters of the model with different numbers of grids.

| Number of Grids | $\xi_{max}$ | $\bar{\xi}$ | rRMSE |
|---|---|---|---|
| 210 (14×15) | $4.51 \times 10^{-2}$ | $6.38 \times 10^{-3}$ | $1.89 \times 10^{-3}$ |
| 840 (28×30) | $\mathbf{1.50 \times 10^{-2}}$ | $\mathbf{3.60 \times 10^{-4}}$ | $\mathbf{1.32 \times 10^{-4}}$ |
| 1890 (42×45) | $1.04 \times 10^{-1}$ | $5.51 \times 10^{-3}$ | $4.91 \times 10^{-4}$ |
| 3660 (56×60) | $3.26 \times 10^{-2}$ | $6.30 \times 10^{-3}$ | $3.93 \times 10^{-4}$ |

## 4.2 Diagnostics with standard deviations

In this section, we applied the IDA model with 840 (28×30) grid points and $k_{jk}^{SE}(\psi_j, \psi_k)$. We introduce different levels of standard deviation—2%, 5%, and 10%—into the synthetic data $d$ and $v_*$ of the virtual diagnostics. The $\sigma_*$ in Eq.(2-5) and the $\Sigma_\varepsilon$ in Eq.(2-11) were adjusted accordingly. The comparison results are shown in **Figure 7**. Column (b) illustrates the comparison between the synthetic diagnostic data and the back-projections (BPs) derived from the reconstructed profiles. The BP is defined as the projection of the 2D reconstructed profile back into the detector's measurement space[41].

As shown in column (a), the reconstructed profiles are generally consistent across the different levels of standard deviation. In column (b), it is observed that the back-projections (BPs, crosses) agree well with the synthetic diagnostic data from FIR. Notably, the BPs are not affected by the varying levels of standard deviation. Column (c) reveals that as the standard deviation increases, the uncertainty of the reconstructed values at each grid point also increases. Computationally, this can be attributed to the fact that as the standard deviation grows, the diagonal elements of $\Sigma^{post}$ increase, leading to larger uncertainties in the results. Physically, a higher standard deviation in the measurement values introduces greater uncertainty into the inference process of the Bayesian model.

(a) (b) (c)

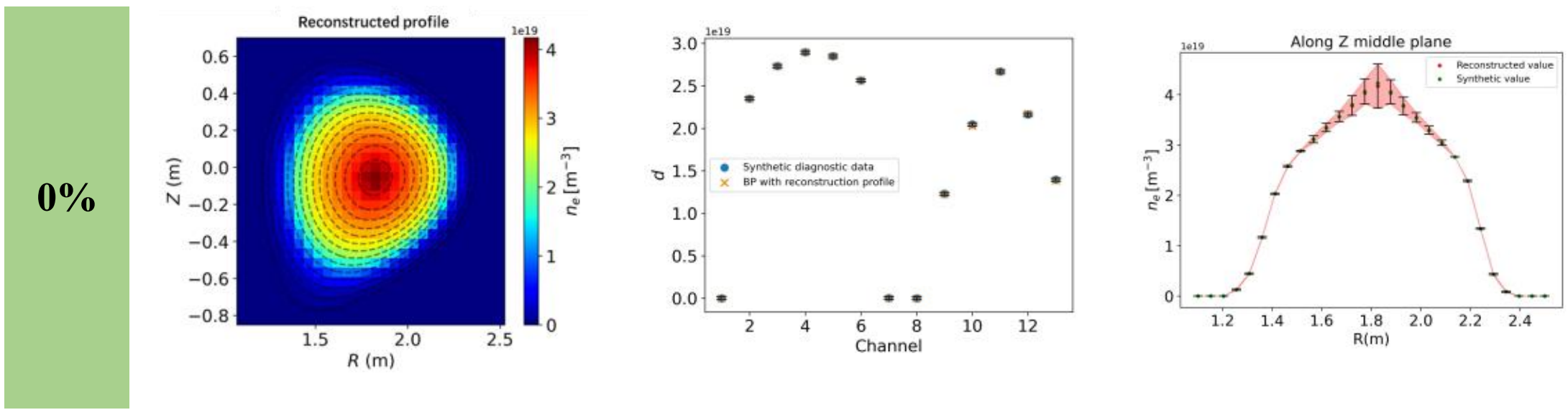

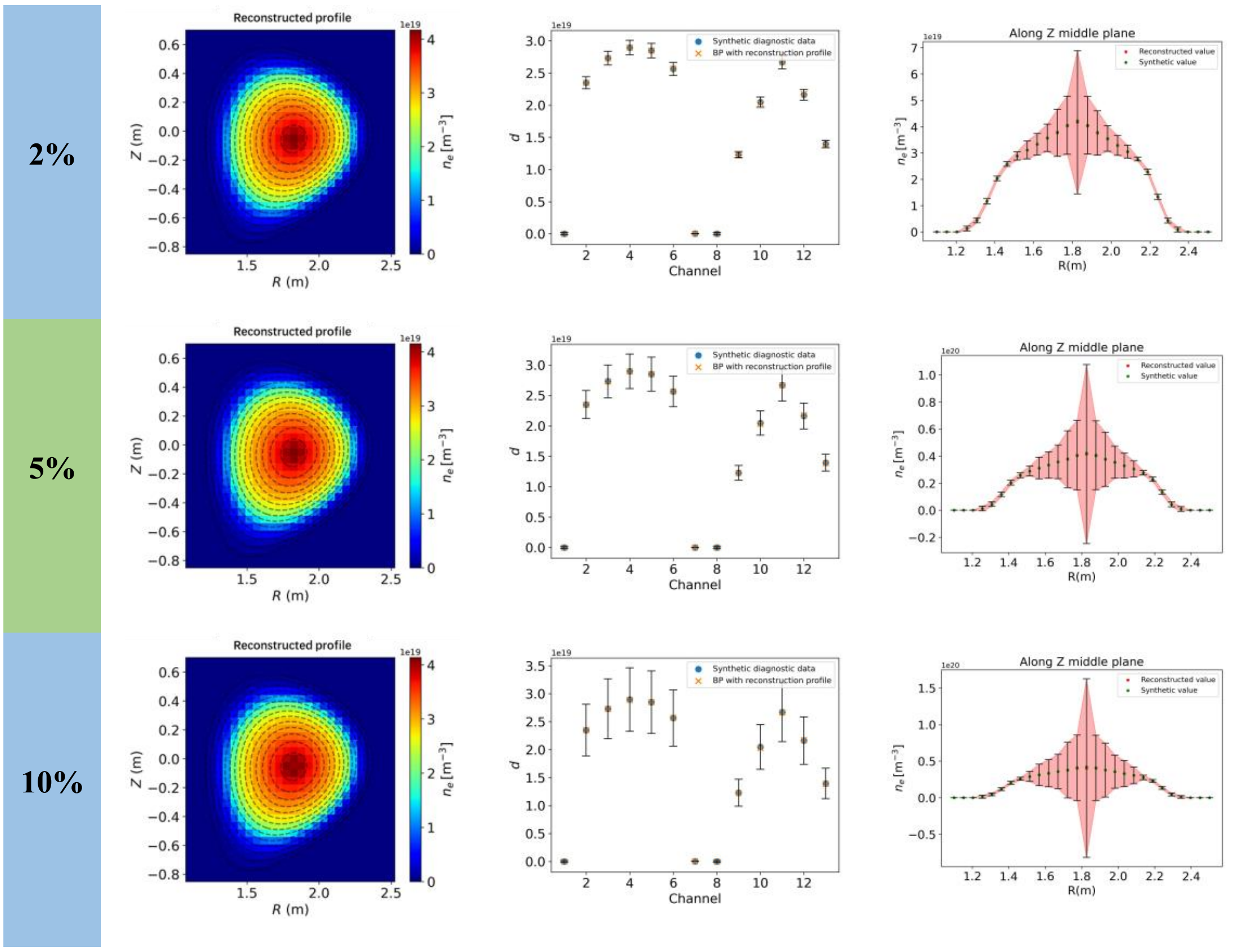

**Figure 7** Inversion results of the IDA model with different standard deviations. Column (a): the reconstructed profiles. Column (b): the comparison between the back-projections (BPs, crosses), calculated based on the reconstructed profile in column (a), and the synthetic diagnostic data from FIR with standard deviation. Column (c): the plasma electron density along the Z midplane of the reconstructed profile.

**Table 4** presents the evaluation parameters in four cases. The $\bar{\xi}$ is approximately $5.10 \times 10^{-4}$ for the IDA model across different standard deviations of the synthetic diagnostic data. The rRMSE increases as the standard deviation increases, reaching a maximum of $2.08 \times 10^{-4}$. Therefore, the IDA model is considered to be robust to variations in the standard deviation of the diagnostic data and can still provide reliable inversion results even at a 10% standard deviation.

**Table 4** Comparison of evaluation parameters of the model with different standard deviations.

| Standard deviation | $\xi_{max}$ | $\bar{\xi}$ | rRMSE |
|---|---|---|---|
| 0% | $1.50 \times 10^{-2}$ | $3.60 \times 10^{-4}$ | $1.32 \times 10^{-4}$ |
| 2% | $1.30 \times 10^{-2}$ | $5.24 \times 10^{-4}$ | $1.33 \times 10^{-4}$ |
| 5% | $1.83 \times 10^{-2}$ | $5.18 \times 10^{-4}$ | $1.40 \times 10^{-4}$ |
| 10% | $4.48 \times 10^{-2}$ | $6.50 \times 10^{-4}$ | $2.08 \times 10^{-4}$ |

## 4.3 Diagnostics with noise

To evaluate the stability of the IDA model, 1000 sets of diagnostic data with varying levels of random noise (1%, 3%, 5%) generated from a Gaussian distribution are applied. The standard deviation for the virtual diagnostics of the synthetic data $d$ and $v_*$ is set to be 2%. According to the optimization criterion of Bayesian Occam's razor discussed in Appendix A, the optimal values for the two hyper-parameters $\sigma$ and $l$ are fixed.

**Table 5** presents the mean of the $\xi_{max}$, $\bar{\xi}$, rRMSE across 1000 samples under three distinct noise conditions: 1%, 3%, and 5%. For reference, the results without added noise were obtained in **Section 4.2**. As the noise level increases from 0% to 5%, the mean values of these evaluation parameters also rise. Specifically, the mean of $\xi_{max}$ increases from $1.30 \times 10^{-2}$ to $6.55 \times 10^{-1}$, the mean of $\bar{\xi}$ grows from $5.24 \times 10^{-4}$ to $1.56 \times 10^{-2}$ and the mean of rRMSE rises from $1.33 \times 10^{-4}$ to $5.13 \times 10^{-3}$. The observed error escalation - with $\xi_{max}$ and rRMSE increasing tenfold (10×) and $\bar{\xi}$ increasing hundredfold (100×) - quantitatively demonstrates the model's pronounced noise sensitivity. These findings provide critical reliability benchmarks for practical implementation scenarios where noise is unavoidable.

**Table 5** Mean of evaluation parameters of the model with different random noises.

| Random noise | Mean($\xi_{max}$) | Mean($\bar{\xi}$) | Mean(rRMSE) |
|---|---|---|---|
| Noise free | $1.30 \times 10^{-2}$ | $5.24 \times 10^{-4}$ | $1.33 \times 10^{-4}$ |
| 1% | $1.42 \times 10^{-1}$ | $3.23 \times 10^{-3}$ | $1.05 \times 10^{-3}$ |
| 3% | $3.96 \times 10^{-1}$ | $9.38 \times 10^{-3}$ | $3.09 \times 10^{-3}$ |
| 5% | $6.55 \times 10^{-1}$ | $1.56 \times 10^{-2}$ | $5.13 \times 10^{-3}$ |

**Table 6** and **Table 7** present statistical summaries, including the median and standard deviation of $\xi_{max}$, $\bar{\xi}$, rRMSE across 1000 samples, to elucidate the central tendency and dispersion of the evaluation parameters at each noise level. **Figure 8** illustrates the frequency distribution histogram and value distribution of the evaluation parameters under 1%, 3%, and 5% noise conditions. For $\xi_{max}$, the median is less than the mean, indicating a positively skewed distribution, as shown in column (a) of **Figure 8**. A larger difference between the mean and median suggests a more pronounced skew. In contrast, for $\bar{\xi}$ and rRMSE, the median is close to the mean, indicating an approximately normal distribution, as shown in columns (b) and (c) of **Figure 8**. Additionally, the standard deviation of these evaluation parameters increases with rising noise levels, indicating greater fluctuations and highlighting the increasing impact of noise on model stability.

**Table 6** Median of evaluation parameters of the model with different random noises.

| Random noise | Median($\xi_{max}$) | Median($\bar{\xi}$) | Median(rRMSE) |
|---|---|---|---|
| 1% | $1.23 \times 10^{-1}$ | $3.14 \times 10^{-3}$ | $1.02 \times 10^{-3}$ |
| 3% | $3.28 \times 10^{-1}$ | $9.11 \times 10^{-3}$ | $2.99 \times 10^{-3}$ |
| 5% | $5.73 \times 10^{-1}$ | $1.50 \times 10^{-2}$ | $4.95 \times 10^{-3}$ |

**Table 7** Standard deviation of evaluation parameters of the model with different random noises.

| Random noise | $\sigma(\xi_{max})$ | $\sigma(\bar{\xi})$ | $\sigma(r\text{RMSE})$ |
|---|---|---|---|
| 1% | $8.19 \times 10^{-2}$ | $9.43 \times 10^{-4}$ | $3.37 \times 10^{-4}$ |
| 3% | $2.36 \times 10^{-1}$ | $2.86 \times 10^{-3}$ | $1.01 \times 10^{-3}$ |
| 5% | $3.58 \times 10^{-1}$ | $4.81 \times 10^{-3}$ | $1.66 \times 10^{-3}$ |

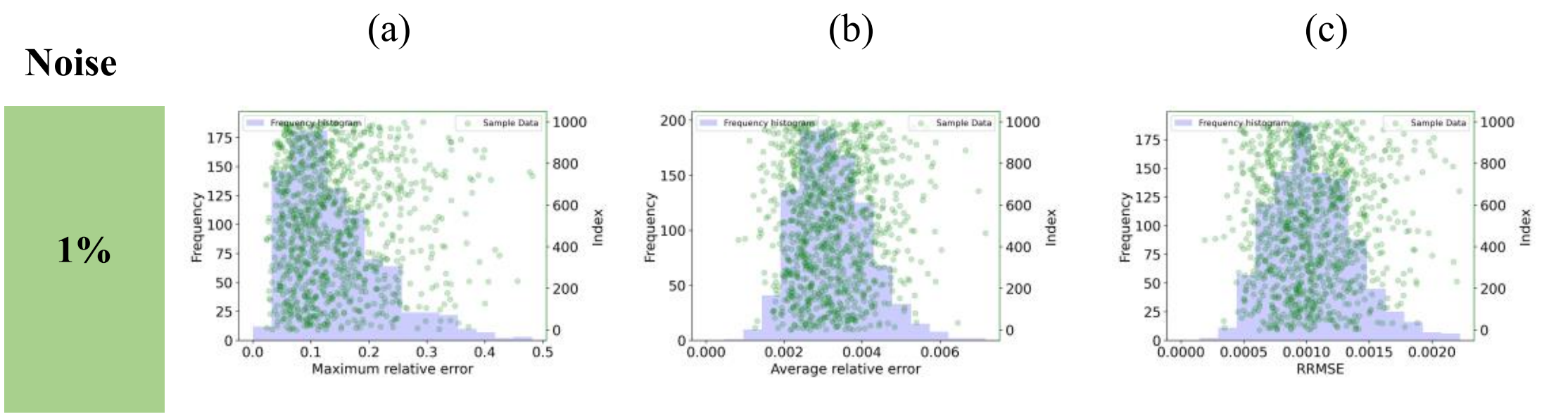

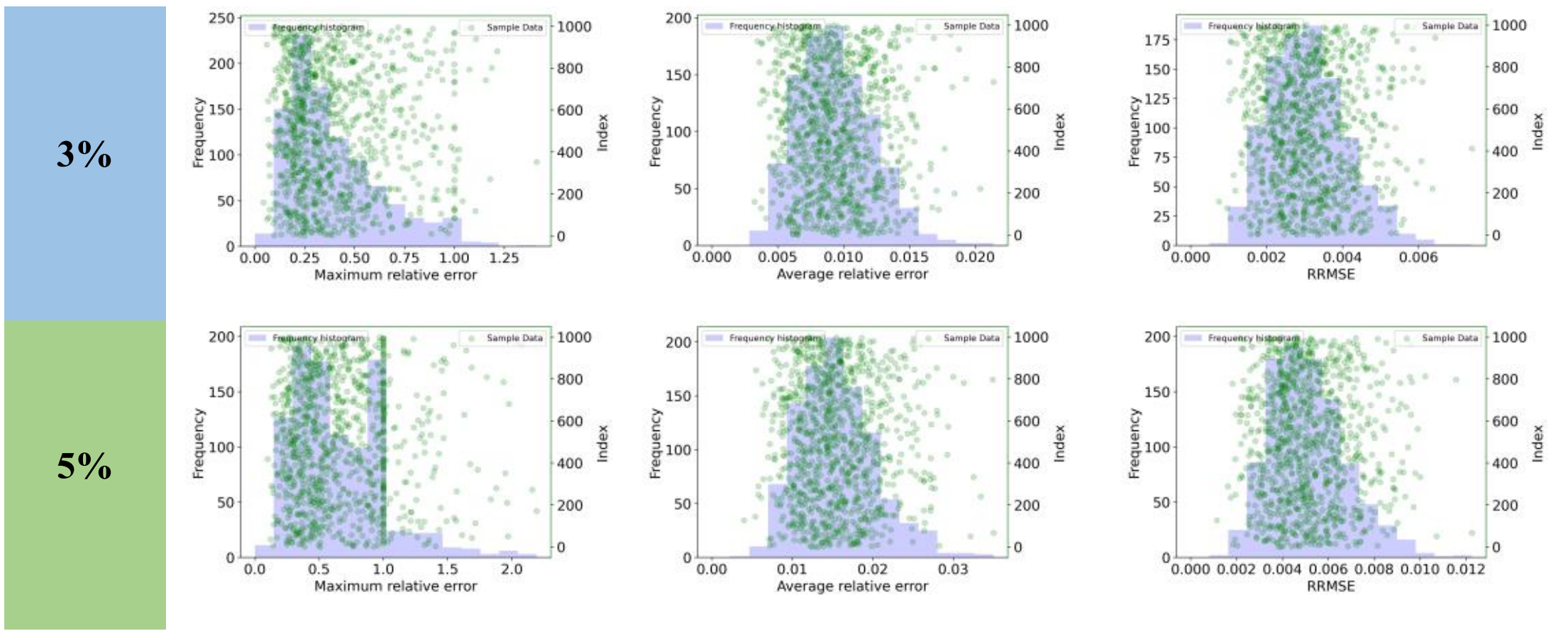

**Figure 8** The blue vertical stripes represent the frequency distribution histogram of the evaluation parameters, while the green points represent the distribution of values. Column (a): the frequency and value distribution of the maximum relative error. Column (b): the frequency and value distribution of the average relative error. Column (c): the frequency and value distribution of the rRMSE.

# 5 Conclusion

In this study, an integrated data analysis model based on Gaussian Process Regression for plasma electron density profile tomography is presented for the HL-3. The results of the IDA model using $k_{jk}^{SE}(x_j, x_k)$ indicate that without magnetic equilibrium information, the current FIR and FMCW measurements alone cannot accurately reconstruct values in the core of the 2D plasma electron density profile. Therefore, it is necessary to either increase the measurement channels in the plasma core or combine with results from other diagnostic systems observing the plasma core. Statistical analysis reveals that adopting normalized magnetic flux coordinates leads to marked error reduction: (1) the $\xi_{max}$ diminishes by 87.5% to $1.50 \times 10^{-2}$; (2) the $\bar{\xi}$ shows a 93.1% improvement, reaching $3.60 \times 10^{-4}$; and the rRMSE decreases to $1.32 \times 10^{-4}$, representing a 91.5% enhancement in overall accuracy. These improvements reflect the enhancement brought by accurate normalized magnetic flux and demonstrate the robustness and scalability of the IDA in this work. Furthermore, we performed systematic sensitivity analyses to evaluate the model's performance under various conditions, including grid resolution, standard deviation of diagnostic data, and noise level. The IDA model demonstrates optimal reconstruction accuracy when employing an 840-point (28×30) grid configuration. Remarkably, the model maintains robust inversion capability even when subjected to 10% measurement uncertainty. Our noise sensitivity analysis establishes quantitative reliability thresholds that are particularly valuable for practical applications involving varying noise conditions. These systematic investigations provide crucial validation for implementing the model in HL-3 experimental data analysis, while clearly delineating its operational boundaries under different noise scenarios.

In future work, the model will be applied to experimental data. A non-stationary kernel function will replace the current kernel function to achieve better results. Furthermore, more point diagnostics and line-integrated diagnostics will be incorporated into the IDA model to improve the inversion accuracy of the 2D profile. Systematic errors arising from EFIT equilibrium reconstruction will be further considered.

## Acknowledgement

*The first author extends heartfelt gratitude to Dr. Haoxi Wang and Dr. Zengchen Yang for their insightful discussions and invaluable contributions. This work is supported by National Natural Science Foundation of China (No.12075078), National Natural Science Foundation of China (No.52406198) and National Natural Science Foundation of China (No. 12405266).*

## Appendix

## A. Bayesian probability theory

In the context of the fusion diagnostic inversion problem, the principles of Bayes' rule can be expressed in a specific form, as illustrated in Eq.(5-1). This formulation facilitates the updating of the prior distribution to the posterior distribution by integrating the information provided by observation data:

$$\begin{aligned} p(F|d;\theta) &= \frac{p(d|F;\theta)\cdot p(F|\theta)}{p(d|\theta)} \\ &\propto p(d|F;\theta)\cdot p(F|\theta), \end{aligned} \tag{5-1}$$

where $F$ represents the profile parameters of interest, such as $T_e$, $n_e$ et al., $d$ denotes the measured data obtained from the observation of $F$, and $\theta$ encompasses the prior information of the probabilistic model, e.g., all the model parameters in the calculation[42]. The term $p(d|\theta)$ serves as a normalization factor, often referred to as the model evidence or marginal likelihood. It quantifies the overall probability of observing the data under the given model assumptions and is computed as:

$$p(d|\theta) = \int p(d|F;\theta)p(F|\theta)dF. \tag{5-2}$$

Within the context of a given model $p(d|\theta)$ does not affect the conclusions. However, according to the optimization criterion of Bayesian Occam's razor, the optimal values of the two hyper-parameters $\sigma$ and $l$ can be determined by maximizing the evidence term $p(d|\theta)$[32,38].

The prior probability distribution, denoted as $p(F|\theta)$, characterizes the range of possible values that $F$ can take based on existing prior knowledge. It encapsulates our initial beliefs or assumptions about $F$ before conducting any experiments. The predictive distribution of the observations $d$, given $F$, is represented by $p(d|F;\theta)$, also referred to as the likelihood function of $F$. The mean of the predictive distribution is typically determined by the forward model, as previously discussed.

The posterior probability distribution, denoted as $p(F|d;\theta)$, is proportional to the product of the prior distribution and the likelihood function. This relation enables the integration of information from both our prior knowledge and the measured data, providing an updated and refined understanding of $F$.

## B. Diagnostics of HL-3 Tokamak

In this study, the frequency modulated continuous wave reflectometry (FMCW) diagnostic system and far-infrared laser interferometer (FIR) diagnostic system of the HL-3 Tokamak are utilized for realizing the Bayesian integrated tomography of electron density inside the Last Closed Flux Surface (LCFS).

### a) HL-3 Tokamak

HL-3 is a medium-sized copper-conductor tokamak [43] located at the Southwestern Institute of Physics (SWIP) in Chengdu, China. It is a totally new machine, as shown in **Figure 9**, with some systems upgraded from the HL-2A tokamak that had been in operation since 2002. HL-3 is designed to have 3MA plasma current, and over 8.6 keV ion temperature. Two of its key missions are to achieve 10 keV ion temperature and investigate the behavior of energetic particles relevant to burning plasmas. With a flexible divertor, a new set of toroidal field coils, and a shaped plasma with improved stability, HL-3 will contribute to establishing the scientific and technical basis for optimizing the tokamak approach to fusion energy and prepare important scaling information for ITER operation.

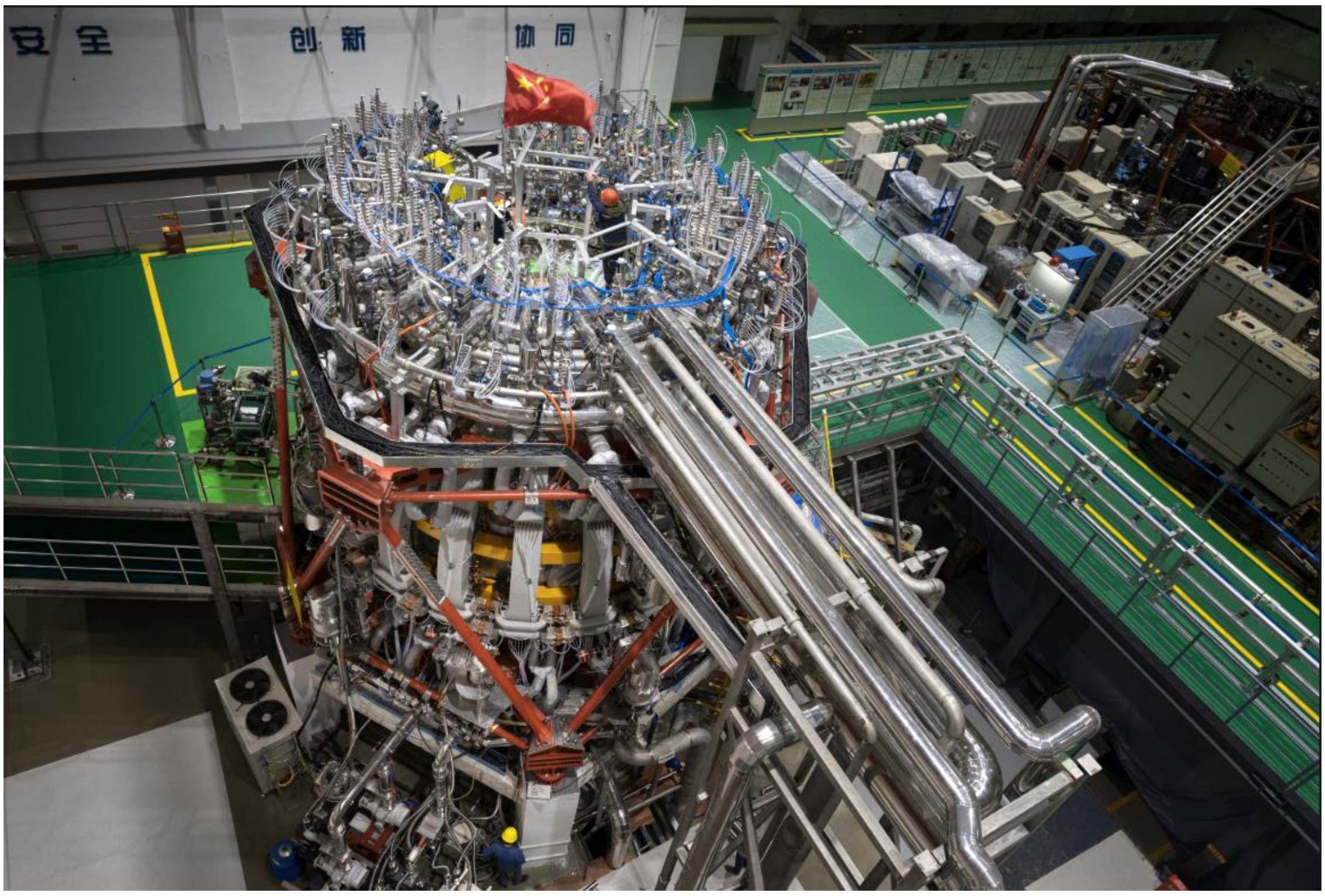

**Figure 9** HL-3 Tokamak.

### b) FMCW Diagnostic System

The FMCW diagnostic system of HL-3 ,which is similar with that of HL-2A[16,44], is employed to measure the local electron density within a region that spans from the outermost LCFS up to approximately one-third of the minor radius at the low-field side, corresponding to magnetic flux coordinates ranging between 0.6 to 1.0. The detection zone is depicted (green area) in **Figure 10**. This particular area is characteristically marked by a significant density gradient, where the density increases monotonically from nearly zero at the LCFS to substantial values in the inner region.

### c) FIR Diagnostic System

The FIR Diagnostic System of HL-3 is designed with 13 detection channels, as shown in **Figure 10**. Among these, there are 8 horizontal channels and 5 oblique channels. The Z coordinates for the horizontal channels are respectively: 0.765m, 0.200m, 0.100m, 0m, -0.100m, -0.200m, -0.760m, and -0.970m, all at an angle of 0° relative to the horizontal plane.

For the oblique channels, their radial coordinate (R) is set at 1.050m, while the Z coordinates are 0.724m, 0.615m, -0.518m, -0.626m, and -0.733m, each inclined at angles of -23°, -23°, 21.5°, 21.5°, and 21.5° respectively. In this work, the synthetic data from the 8 horizontal channels and 5 oblique channels are employed as inputs for line-integral measurements. For the top channel and the two bottom channels in **Figure 10**, there are used to measure the electron density signal at the divertor.

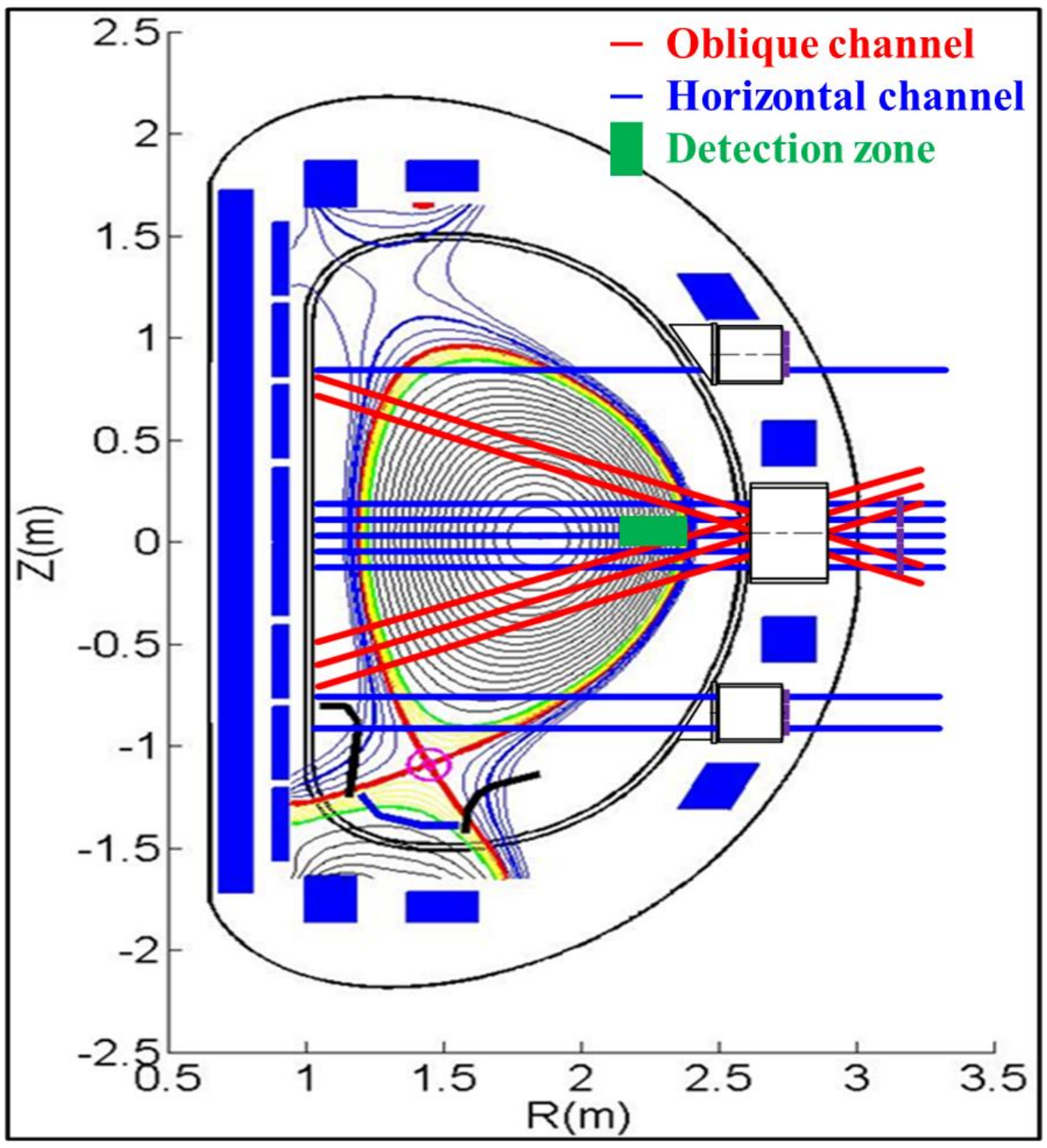

**Figure 10** Poloidal view of the typical plasma configuration on HL-3 with two primary density diagnostics mapped to the same cross-section (R is major radius; Z is the axis of axisymmetry located at R = 0 m): detection zone of the FMCW diagnostic system (green area) and 13 detection channels of FIR diagnostic system (blue lines and red lines).

## C. Synthetic profile

The frequency modulated continuous wave reflectometry (FMCW) diagnostic system and far-infrared laser interferometer (FIR) diagnostic system of the HL-3 Tokamak are briefly introduced in Appendix B. To build the synthetic profile for HL-3, EFIT is employed to design a self-consistent equilibrium magnetic flux, as depicted in **Figure 11**. The electron density profile is subsequently modeled using a modified tanhfit function[45] as shown in Eq.(5-3). $\rho$ represents the normalized toroidal magnetic flux. XSYM denotes the location of the center of the barrier, while HWID refers to the half-width of the barrier. $\alpha$ enables a smooth transition to a linear fit near the core profile. Parameters A and B are utilized to adjust the magnitude and minimum value of the electron density.

$$\bar{n}_e(\rho) = A * MTANH(\alpha, z) + B \tag{5-3}$$

$$MTANH(\alpha, z) = \frac{(1 + \alpha \cdot z)\exp(z) - \exp(-z)}{\exp(z) + \exp(-z)}$$
$$z = \frac{XSYM - \rho}{HWID} \tag{5-4}$$

**Figure 12**(a) illustrates the distribution of electron density along the normalized magnetic plane, while **Figure 12**(b) demonstrates the 2D electron density profile. The line-integral measurements of FIR diagnostic system $d$ are obtained according to the positions of 13 channels and the electron density profile.

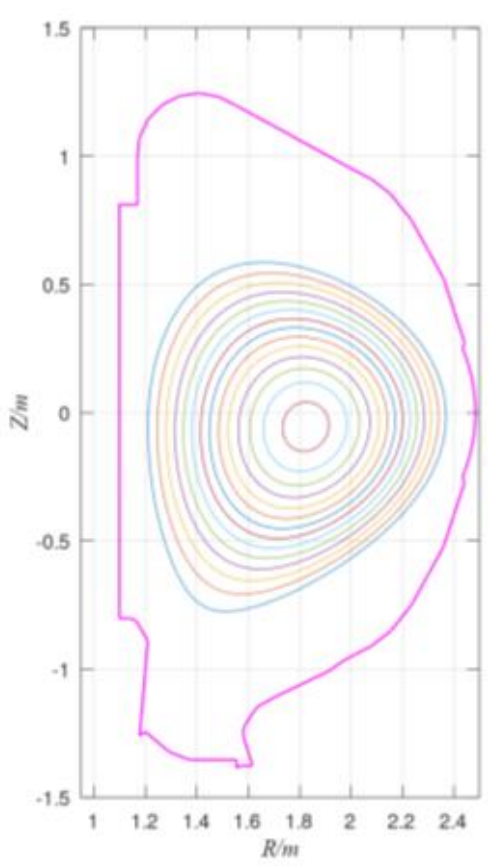

**Figure 11** A self-consistent equilibrium magnetic flux.

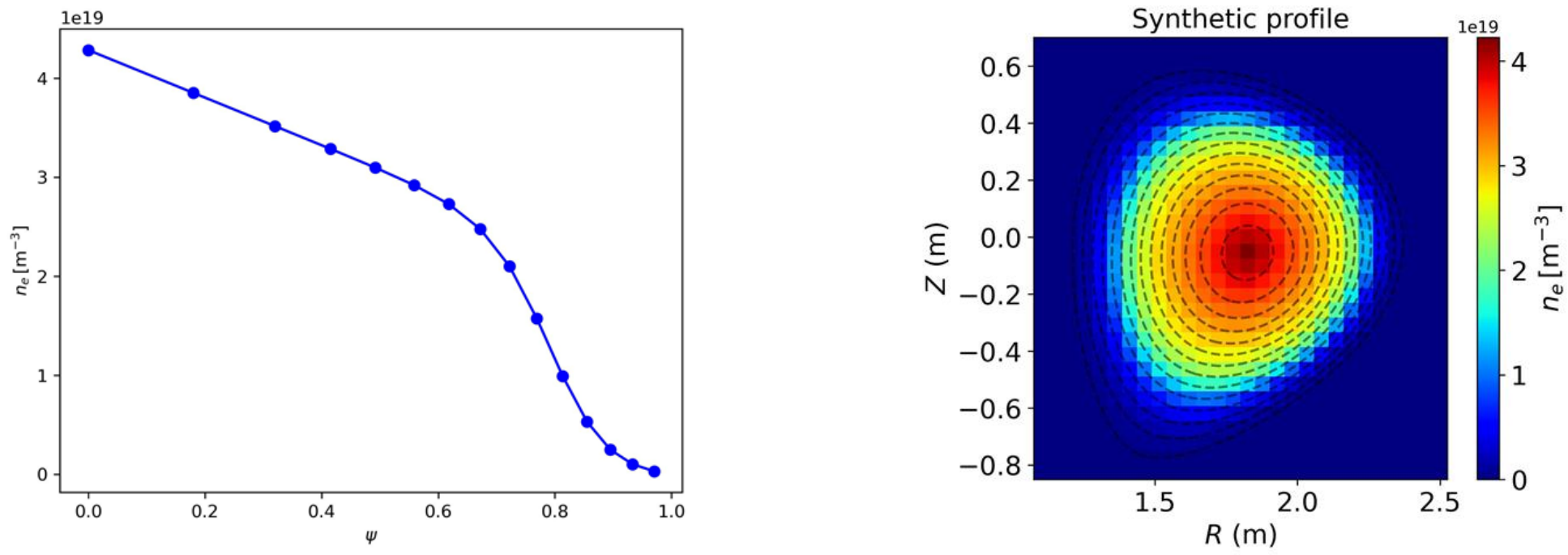

(a) (b)

**Figure 12** Synthetic profile of electron density. (a) shows the distribution of electron density along the normalized magnetic surface; (b) shows the 2D synthetic profile of electron density